\documentclass[11pt]{article}
\usepackage{amsmath,amssymb}
\usepackage{graphicx}
\usepackage{booktabs}
\usepackage{filecontents}
\title{Knowledge Graph and Hypergraph Transformers with Repository-Attention and Journey-Based Role Transport}
\author{Mahesh Godavarti}
\date{}
\begin{document}
\maketitle

\begin{abstract}
We present a concise architecture for joint training on sentences and structured data while keeping knowledge and language representations separable. The model treats knowledge graphs and hypergraphs as structured instances with role slots and encodes them into a key--value repository that a language transformer can attend over. Attention is conditioned by journey-based role transport, which unifies edge-labeled KG traversal, hyperedge traversal, and sentence structure. We outline a dual-stream architecture, hierarchical layer groups with instance-local, neighborhood, and global mixing attention, retrieval over a separate repository, and multi-task objectives spanning masked language modeling, link prediction, and role-consistency denoising. The result is an explicit, inspectable separation between linguistic context and structured knowledge, while still enabling tight alignment through cross-attention.
\end{abstract}

\section{Introduction}
Transformers \cite{Vaswani2017} have been extended to graphs and knowledge bases in many forms. Graph transformers such as HGT \cite{Hu2020HGT} and Graphormer \cite{Ying2021Graphormer} add structural biases to attention. KG-focused transformers such as KG-BERT \cite{Yao2019KGBERT} and CoLAKE \cite{Sun2020CoLAKE} treat triples and text as token sequences. Hypergraph models such as HGNN \cite{Feng2018HGNN}, Hyper-SAGNN \cite{Zhang2019HyperSAGNN}, and HyperGT \cite{Liu2024HyperGT} capture higher-order relations that are not reducible to pairwise edges.

A recurring challenge is balancing general language modeling with faithful use of structured knowledge. This paper specifies an architecture that separates knowledge from language via a repository of structured key--value (KV) items, while still enabling joint training and reasoning through role-conditioned attention.

\section{Background}
\subsection{Two views of knowledge graphs}
A KG can be represented as an \emph{edge-labeled multigraph}, where each triple $(h,r,t)$ is a directed edge $h \xrightarrow{r} t$, or as a \emph{reified/incidence graph}, where each relation instance is a node (or hyperedge) connected to its participants. These views are equivalent but lead to different attention formulations.

\subsection{Hypergraphs}
Hypergraphs generalize graphs by allowing a hyperedge to connect any number of nodes. Many facts are naturally n-ary (predicate, arguments, time, location, source, etc.), so representing them as hyperedges preserves role integrity.

\subsection{Sentence--graph equivalence}
A sentence can be interpreted as:
\begin{enumerate}
  \item a sequence of tokens;
  \item an edge-labeled KG with relation \textsc{next} between adjacent tokens;
  \item a semantic-role hyperedge (PREDICATE, ARG1, ARG2, TIME, ...);
  \item a position-slot hyperedge with slots POSITION$_k$ for $k=1..T$;
  \item a POS-slot hyperedge (NOUN, VERB, ...) with within-slot positions.
\end{enumerate}
This equivalence enables a single attention mechanism to operate over language and structured facts by changing only the slot schema.
In practice, a sentence can appear as multiple structured instances in the same batch: a position-slot hyperedge and a POS-slot hyperedge. The same token can be present in both views (shared value embedding or a linking micro-hyperedge), allowing attention to connect a token to itself across views and combine position-based and POS-based journeys.

\section{Journey-Based Role Transport}
JoFormer \cite{Godavarti2025JoFormer} encodes relative position as a composed journey operator. We extend this to roles and relation instances. Each token has a slot label $s(i)$ with a learned operator $R_{s(i)}$. A journey from role $a$ to role $b$ is
\[
P_{a\to b} = R_a R_b^{-1}.
\]
Attention between tokens $i$ and $j$ is then
\[
\text{score}(i,j) = \frac{q_i^\top P_{s(i)\to s(j)} k_j}{\sqrt{d}} + b_{s(i),s(j)}.
\]
This yields direct edge traversal in the edge-labeled view and internal-node traversal in the reified/hyperedge view.

\paragraph{Recovering RoPE.}
When slot labels correspond to absolute positions, the journey operator recovers rotary positional embeddings (RoPE) \cite{Su2021RoFormer}. Let $s(i)=\text{POSITION}_i$ and choose orthogonal, block-diagonal operators
\[
R_i = \operatorname{diag}(R(\theta_1 i), \ldots, R(\theta_{d/2} i)),
\]
with each $R(\cdot)$ a $2\times2$ rotation. Then
\[
P_{i\to j} = R_i R_j^{-1} = R_{i-j},
\]
and
\[
q_i^\top P_{i\to j} k_j = (R_i q_i)^\top (R_j k_j),
\]
which is exactly the RoPE dot-product form. Thus RoPE is a special case of role transport, and the same mechanism extends from positions to arbitrary roles and instances.

\subsection{Edge-labeled multigraph journeys}
In the edge-labeled view, a relation type $r$ is a directed transport operator $R_r$. A journey from head $h$ to tail $t$ along relation $r$ is
\[
J_{h\to t} = R_r, \quad J_{t\to h} = R_r^{-1}.
\]
A simple attention score is
\[
\text{score}(h,t) = \frac{q_h^\top R_r k_t}{\sqrt{d}} + b_r,
\]
which yields role-aware transport on edges without introducing internal nodes.

\subsection{Hyperedge and reified journeys}
In a reified or hypergraph view, each fact instance has an internal node (or hyperedge) that mediates interactions among participants. With role operators $R_a$ and $R_b$, the journey from role $a$ to role $b$ via the instance is
\[
P_{a\to b} = R_a R_b^{-1}.
\]
Optional within-slot position operators can be composed at the endpoints when a role is represented by multiple tokens.

\section{Repository-Attention Architecture}
\subsection{Dual-stream design}
We use two streams:
\begin{enumerate}
  \item \textbf{Language stream} for sentence tokens (sequence or sentence hyperedges).
  \item \textbf{Structured stream} for KG triples and hypergraph facts.
\end{enumerate}
Structured instances are encoded into a separate repository of KV items that the language stream can attend over.

\subsection{Hierarchical receptive fields}
Within each stream, layers can be grouped by receptive field:
\begin{enumerate}
  \item \textbf{Instance-local layers} attend only within each structured instance to preserve role integrity.
  \item \textbf{Neighborhood layers} attend across linked instances (e.g., shared entities or shared sentence tokens).
  \item \textbf{Global mixing layers} attend over the provided set or retrieved repository items. These layers may omit explicit positional encodings, relying on contextualized representations from lower layers.
\end{enumerate}

\subsection{Structured KV representation}
Each structured instance yields a small set of tokens with slot labels (HEAD, RELATION, TAIL, ARG$_k$, TIME, etc.). For a token $x_j$ in slot $s(j)$ we compute
\[
 k_j = W_k R_{s(j)} x_j, \quad v_j = W_v x_j,
\]
and store $(k_j, v_j, s(j), e(j))$ in the repository, where $e(j)$ is the instance id. Instance operators $R_{e(j)}$ can be composed with slot operators when cross-instance transport is desired.

\subsection{Cross-attention to the repository}
For a language token $i$ (with slot label WORD or POS), the query $q_i$ attends to retrieved repository items:
\[
\alpha_{ij} \propto \exp\left(\frac{q_i^\top R_{s(i)} R_{s(j)}^{-1} k_j}{\sqrt{d}}\right),
\quad y_i = \sum_j \alpha_{ij} v_j.
\]
This allows attention to connect a token to itself across views (e.g., via POSITION$_k$ and via POS slots) as well as to KG and hypergraph facts.

\subsection{Position-agnostic repository attention}
Repository attention can be made position-agnostic by omitting explicit positional encodings in the cross-attention block. Lower language layers still encode local order, but the repository lookup depends only on contextualized content and role transport:
\[
\alpha_{ij} \propto \exp\left(\frac{q_i^\top P_{s(i)\to s(j)} k_j}{\sqrt{d}}\right),
\quad P_{s(i)\to s(j)} = R_{s(i)} R_{s(j)}^{-1}.
\]
This yields content-driven retrieval that is robust to sentence length or position shifts while preserving slot integrity.

\subsection{Hyperedge-instance cross attention}
For hyperedges or KG facts, each instance $e$ may have an operator $R_e$ that encodes instance identity or provenance. Cross-attention from a source token in instance $e_1$ to a repository token in instance $e_2$ uses an instance-aware journey:
\[
P_{(s,e_1)\to(s',e_2)} = R_{s} R_{e_1} R_{e_2}^{-1} R_{s'}^{-1},
\]
so that instance boundaries are explicit in attention. This enables selective mixing within a fact, across related facts, or between retrieved and input instances.
Long-context interactions can be expressed via instance journeys (e.g., $R_{e_1} R_{e_2}^{-1}$ or multi-hop compositions across a chain of instances), rather than relying on positional distance in a concatenated sequence. For sentence instances, a cross-sentence interaction between token position $i$ in sentence $e_1$ and position $j$ in sentence $e_2$ can compose within-sentence positions with the instance journey:
\[
P_{(i,e_1)\to(j,e_2)} = R_i R_{e_1} R_{e_2}^{-1} R_j^{-1},
\]
preserving local order within each sentence while enabling long-range binding. Instance operators can be derived from token content via pooling, a learned readout token, or attention pooling followed by an MLP, and may be head-specific or conditioned on slot labels or provenance. Parameterizations such as orthogonal (e.g., block-diagonal $2\times2$ rotations), diagonal, or low-rank operators with stability constraints make inversion and instance-journey transport robust.

\subsection{Separate repository and retrieval}
The repository can be kept external to the language model, enabling explicit separation of knowledge and language. Retrieval (approximate nearest neighbor or learned routing) selects a small set of KV items per batch, similar in spirit to RAG \cite{Lewis2020RAG}, REALM \cite{Guu2020REALM}, or RETRO \cite{Borgeaud2022RETRO}. This reduces context length and makes knowledge updates modular without retraining the language stream.

\section{Joint Training Objectives}
We jointly train on sentences and structured instances while keeping the repository distinct.
\begin{itemize}
  \item \textbf{Masked modeling} on sentence tokens and structured-instance tokens (entities, predicates, or qualifiers).
  \item \textbf{Link prediction / completion} for KG triples and hyperedges.
  \item \textbf{Role-consistency denoising} that swaps qualifiers across instances and trains recovery.
  \item \textbf{Joint text+KG training} mixing sentence and structured instances in the same batch with shared entity tokens.
  \item \textbf{Alignment losses} between sentence spans and entity nodes (contrastive or retrieval loss) to encourage correct repository access.
  \item \textbf{Memory-based objectives} such as kNN-LM style next-token retrieval \cite{Khandelwal2020KNNLM}.
\end{itemize}
This setup permits a clean separation of knowledge storage while still enabling tight cross-attention between language and structure.

\section{Discussion}
Separating knowledge from language clarifies what is stored versus inferred. The repository provides inspectable facts, and the language model provides composition. Journey-based attention unifies positional transport, KG traversal, and hyperedge traversal; in sequences the path follows indices, while in KGs and hypergraphs the path follows roles and instances. The same token can appear in multiple structured instances (sequence, POS hyperedge, position hyperedge), enabling multi-view consistency within a single architecture.

\section{Conclusion}
We presented a repository-attention architecture for KG and hypergraph transformers that explicitly separates structured knowledge from language while supporting joint training. Role-conditioned journey operators provide a unifying attention mechanism, and KV repositories make knowledge modular, updateable, and inspectable.

\begingroup
\raggedright
\bibliographystyle{unsrt}
\bibliography{references2}
\endgroup
\end{document}